\documentclass[10pt,conference]{IEEEtran}

\usepackage{tikz}

\usepackage{amsmath,amsfonts}
\usepackage{amsfonts}
\usepackage{booktabs}
\usepackage{siunitx}
\usepackage{algorithmic}
\usepackage{array}
\usepackage[caption=false,font=normalsize,labelfont=sf,textfont=sf]{subfig}
\usepackage{textcomp}
\usepackage{stfloats}
\usepackage{url}
\usepackage{verbatim}
\usepackage{graphicx}
\usepackage{authblk}
\usepackage{caption}
\usepackage{color}
\usepackage{soul}
\usepackage{orcidlink}

\IEEEoverridecommandlockouts

\def\BibTeX{{\rm B\kern-.05em{\sc i\kern-.025em b}\kern-.08em
    T\kern-.1667em\lower.7ex\hbox{E}\kern-.125emX}}
\usepackage{balance}

\begin{document}
\title{Deep UAV Path Planning with Assured Connectivity in Dense Urban Setting\\
\thanks{This work was supported in part by IITP grant funded by the Korean government (MSIT) under IITP-2024-2020-0-01821(40\%), IITP-2021-0-02068(30\%), and IITP-2019-0-00421(30\%).}
}

\author[1]{Jiyong Oh}
\author[2]{Syed M. Raza \orcidlink{0000-0001-6580-3232}}
\author[3]{Lusungu J. Mwasinga}
\author[4]{Moonseong Kim}
\author[2*]{Hyunseung Choo \orcidlink{0000-0002-6485-3155}}
\affil[1]{{\small Dept. of Artificial Intelligence, Sungkyunkwan University, Suwon, Korea.}}
\affil[2]{{\small Dept. of Electrical and Computer Engineering, Sungkyunkwan University, Suwon, Korea.}}
\affil[3]{{\small Dept. of Computer Science and Engineering, Sungkyunkwan University, Suwon, Korea.}}
\affil[4]{{\small Dept. of IT Convergence Software, Seoul Theological University, Bucheon, Korea.}}
\affil[*]{{\small Corresponding author (choo@skku.edu)}}

\maketitle
\begin{abstract}
Unmanned Ariel Vehicle (UAV) services with 5G connectivity is an emerging field with numerous applications. Operator-controlled UAV flights and manual static flight configurations are major limitations for the wide adoption of scalability of UAV services. Several services depend on excellent UAV connectivity with a cellular network and maintaining it is challenging in predetermined flight paths. This paper addresses these limitations by proposing a Deep Reinforcement Learning (DRL) framework for UAV path planning with assured connectivity (DUPAC). During UAV flight, DUPAC determines the best route from a defined source to the destination in terms of distance and signal quality. The viability and performance of DUPAC are evaluated under simulated real-world urban scenarios using the Unity framework. The results confirm that DUPAC achieves an autonomous UAV flight path similar to base method with only 2\% increment while maintaining an average 9\% better connection quality throughout the flight.
\end{abstract}

\begin{IEEEkeywords}
UAV flight planning, signal quality, Deep Reinforcement Learning (DRL), Unity framework
\end{IEEEkeywords}

\section{Introduction}
Unmanned aerial vehicles (UAVs) are emerging as a new service paradigm in various areas including package delivery, rescue efforts, and agriculture \cite{R1_1}. This trend is expected to expand, encompassing a broader range of services that utilize several UAVs \cite{R1_2}. It is not feasible in terms of service flexibility, scalability, and maintenance to manually configure UAV flight paths or hire a control operator for each UAV \cite{R1_3}. This necessitates autonomous and effective UAV flight path planning and uninterrupted connectivity as they depend on batteries and communications for operation and control, respectively \cite{R1_4}.  Effective flight path planning must account for the surrounding complex 3D environment and obstacles to determine the optimal flight path, whereas uninterrupted connectivity facilitates precise control of UAVs, allowing for real-time flight optimization \cite{R1_5}. However, maintaining an uninterrupted excellent connection conflicts with the ideal minimum flight path, as there is no guarantee of consistent signal strength on the optimal route. Consequently, there is a trade-off between these two critical UAV requirements, making their balanced management a significant challenge.

Existing studies have proposed various techniques for addressing UAV path planning and connectivity issues. The study in \cite{R2} employs graph theory to design an optimal UAV path that minimizes flight distance and further uses dynamic programming for UAV path planning. Dijkstra's algorithm has been employed to optimize the trajectories of cellular-connected UAVs \cite{R4}. These methods face challenges when dealing with dynamic environment changes, as they rely on predefined settings. Recently, Deep Reinforcement Learning (DRL) has shown a capability to handle dynamic environmental changes through continuous learning in complex settings. DRL methods like DQN and multi-step dueling DDQN have been used to optimize UAV flight time and reduce expected communication interruptions in \cite{R5,R6}, respectively. However, these studies use discrete grid points to determine the next position of a UAV during the flight which sacrifices the precise maneuvering of a UAV in a dense urban environment and increases risk of collisions. Moreover, it focuses only on maintaining connectivity at UAV and does not take the quality of connection into account.  These substantial limitations underscore the necessity for the development of a flexible and robust UAV path-planning approach while maintaining low delay and high-quality network connectivity.

This study addresses limitations in existing studies by proposing a Deep Reinforcement Learning Approach for UAV Path Planning with Assured Connectivity (DUPAC). DUPAC is designed to achieve a balance between maintaining stable communication links and optimizing flight efficiency. It achieves excellent and seamless connectivity by executing handovers to the most appropriate Ground Base Stations (GBSs) at opportune moments using Reference Signal Received Power (RSRP). The flight path is determined by measuring the distance to the destination in each step and if the distance has increased compared to the previous step DUPAC adjusts the UAV direction and learns not to repeat the action that caused the distance increment. The evaluation of DUPAC is done in 3D dense urban environment implemented in the Unity framework, where GBSs mimic real-world deployments with coverage area defined using an antenna model \cite{R6}, and a path loss model determines RSRP at the UAV. The preliminary results confirm the efficacy of DUPAC as it achieves 5\% better RSRP on average while keeping flight distance similar to a base method where DRL focuses only on minimizing flight distance. In summary, the main contributions of this study are: 

\begin{itemize}
  \item DUPAC framework that integrates connectivity assurance into UAV path planning, and achieves a delicate balance between stable communication at the UAV and minimum flight distance to the given destination.
  \item A transformation of action space from discrete to continuous, which not only enhances the decision-making capabilities in DRL but also significantly improves the precision and flexibility of UAV movements.
 % \item We optimize the handover decision process to enable the UAV to learn the best timing and locations for handovers, thereby minimizing interruptions and ensuring smooth connectivity.
  \item Implementation of the real-world 3D urban environment which enables DUPAC evaluation under realistic settings and provides a foundation to extend this study and develop complex UAV systems.
  
\end{itemize}
%%%%%%%%%%%%%%%%%%%%%%%%%%%%%%%%%%%%%%%%%%%%%%%%%
                % Section II
%%%%%%%%%%%%%%%%%%%%%%%%%%%%%%%%%%%%%%%%%%%%%%%%%

\section{System Model}

An urban environment model consists of high-density arrangement of variable-height buildings, where a flight for UAV $u$ starts from source to destination position. The flight is mapped in 3D coordinate system and at time $t$ the position of $u$ is $u^{p}_{t}=[x_t,y_t,z_t]$, where the altitude $z_t$ value is in the range $z_t \in [z_{\text min}, z_{\text max}]$ as defined by Federal Aviation Administration (FAA) regulations \cite{R7}. From the source, $u$ determines its 3D flight path with an average speed of 75 Km/h and must reach the destination within time limit $T$ such that ${u^{p}_{t}: 0\leq t\leq T}$. The network in the urban environment is modeled with $M$ GBSs, where each GBS $g_m$ is deployed using data from CellMapper which is a source for global mobile network GBS locations \cite{R11}.

The coverage footprint of the GBSs is determined using the antenna radiation model. Each GBS is divided into three sectors, each equipped with a uniform linear array consisting of eight elements vertically arranged. These elements are designed to provide high directional gain. The radiation pattern of each element $E$ is a combination of horizontal $A_{(E, H)}$and vertical $A_{(E, V)}$ radiation patterns. By integrating the vertical and horizontal radiation patterns, we can derive the 3D antenna element gain \(A_E (\theta, \phi)\) for each pair of angles.
\begin{equation}
A_E (\theta, \phi) = G_{E,\text{max}} - \min\left\{-\left[A_{(E,V)} (\theta) + A_{(E,H)} (\phi)\right], A_m\right\}
\label{eq:1}
\end{equation}
The maximum directional gain $G_{E,\text{max}}$ of each antenna element is \(8 \, \text{dBi}\). This provides the dB increase experienced by a beam with a pair of angles \((\theta, \phi)\) due to the effect of the element radiation pattern. A comprehensive antenna array radiation pattern \( A_A (\theta, \phi) \) is represented as a combination of the single-element \(A_E (\theta, \phi)\) radiation pattern and array factor $AF(\theta, \phi, n)$ \cite{R6}.
\begin{equation}
A_A (\theta, \phi) = A_E (\theta, \phi) + AF(\theta, \phi, n),
\label{eq:2}
\end{equation}
where $n$ is the number of antenna elements. Using (\ref{eq:2}), total antenna gain at a location is calculated for the UAV.

\begin{figure}[!t]
\centering
\includegraphics[width=0.375\textwidth]{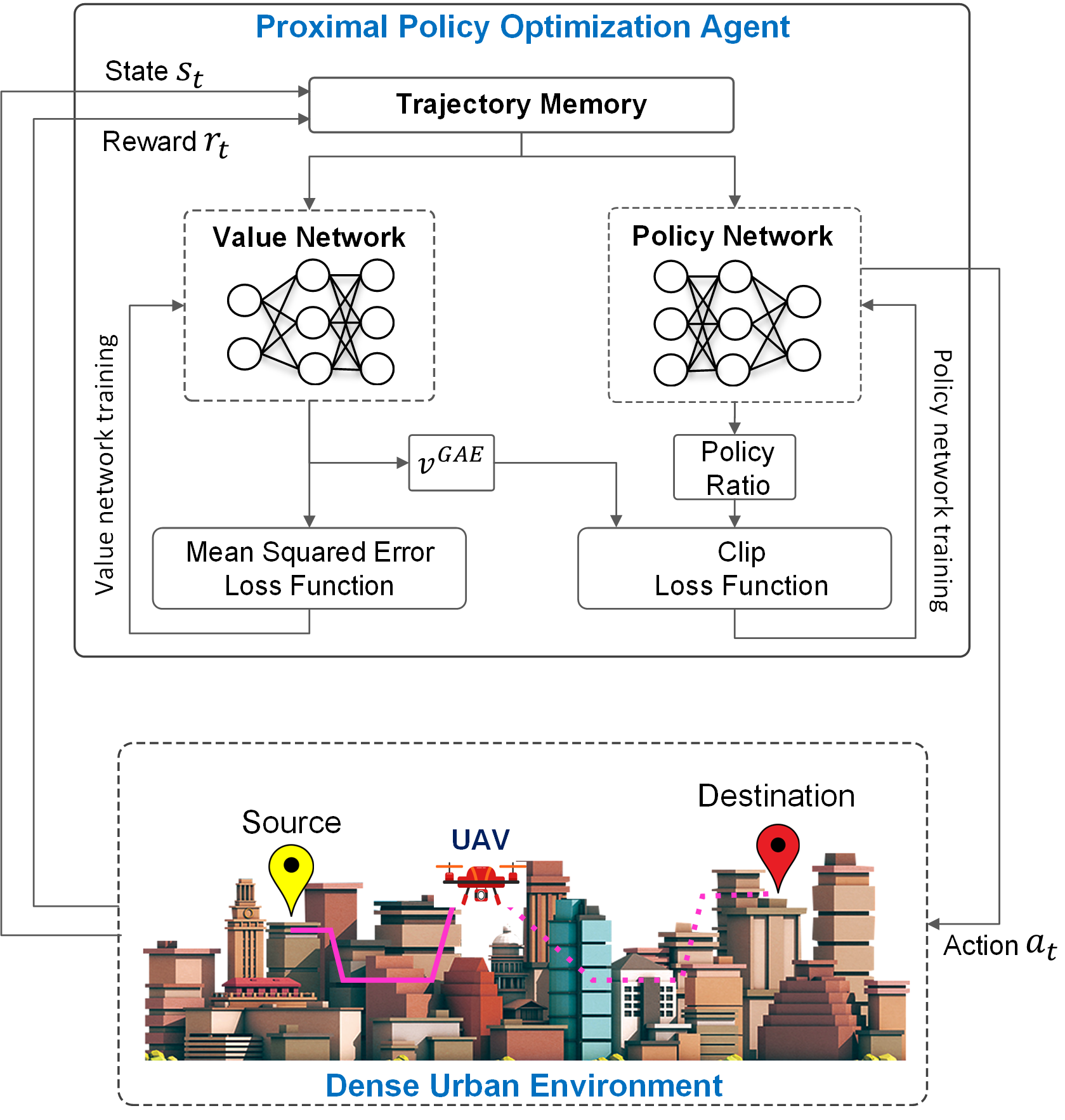}
\caption{Proximal Policy Optimization (PPO) framework for the proposed UAV Path Planning with Assured Connectivity.}
\label{fig_1}
\end{figure}

The RSRP at the UAV is calculated using path loss model that consists of Urban Micro $\omega_{\text{micro}}$ and Urban Macro $\omega_{\text{macro}}$ models specified by 3GPP \cite{R9}. In $\omega_{\text{micro}}$, GBS antenna is installed below the rooftop height, whereas in $\omega_{\text{macro}}$ GBS antenna is installed above the rooftop height of the surrounding buildings. The RSRP at $u$ from GBS $g_m$ at time $t$ is a function of position $u^{p}_{t}$ and is calculated as:
\begin{equation}
\text{RSRP}_{g_m} (u^{p}_{t}) = P_{g_m} + G_{g_m} (u^{p}_{t}) - L_{g_{m}} (u^{p}_{t}), 
\end{equation}
where $P_{g_{m}}$ is the constant reference signal power. $G_{g_m} (u^{p}_{t})$ and $L_{g_{m}} (u^{p}_{t})$ are GBS antenna gain and the path loss, respectively, and they are functions of location $u^{p}_{t}$. The path loss $L_{g_{m}}(u^{p}_{t})$ is calculated for either line-of-sight (LoS) or non-line-of-sight (NLoS) link between GBS $g_m$ and UAV position $u^{p}_{t}$ at time $t$ as follows:
\vspace{-0.1cm}
\begin{equation}
    L_{g_{m}}^{\text{LoS}} (u^{p}_{t})= \begin{cases}
\omega_{1} & \text{if } \omega_{\text{micro}}, 1.5 \leq z_t \leq 22.5 \\
\max\left\{ L_{g_{m}}^{\text{FSPL}}, \omega_{2}\right\} & \text{if } \omega_{\text{micro}}, 22.5 < z_t \leq 300 \\
\omega_{3} & \text{if } \omega_{\text{macro}}, 1.5 \leq z_t \leq 22.5 \\
\omega_{4} & \text{if } \omega_{\text{macro}}, 22.5 < z_t \leq 300
\end{cases}
\end{equation}

where $ L_{g_{m}}^{\text{FSPL}}$ is a free-space path loss between $u^{p}_{t}$ and $g_m$, $\omega_1 = 32.4 + 21 \log d_{g_m}(u^{p}_{t}) + 20 \log f_c$, $\omega_2 = 30.9 + (22.25 - 0.5 \log z_t) \log d_{g_m} (u^{p}_{t}) + 20 \log f_c$, $\omega_3 = 32.4 + 20 \log d_{g_m}(u^{p}_{t}) + 20 \log f_c$, and $\omega_4 = 28 + 22 \log d_{g_m} (u^{p}_{t}) + 20 \log f_c$.  Here, $f_c$ is the frequency, and 3D distance between $u^{p}_{t}$ and $g_m$ at $t$ is represented by $d_{g_m}(u^{p}_{t})$ .

\vspace{-0.4cm}
\begin{equation}
L_{g_{m}}^{\text{NLoS}} (u^{p}_{t})= \begin{cases}
\max\left\{ L_{g_{m}}^{\text{LoS}},\omega_{5}\right\} & \text{if } \omega_{\text{micro}}, 1.5 \leq z_t \leq 22.5 \\
\max\left\{ L_{g_{m}}^{\text{LoS}},\omega_{6}\right\} & \text{if } \omega_{\text{micro}}, 22.5 < z_t \leq 300 \\
\max\left\{ L_{g_{m}}^{\text{LoS}},\omega_{7}\right\} & \text{if } \omega_{\text{macro}}, 1.5 \leq z_t \leq 22.5 \\
\left\{ \omega_{8}\right\} & \text{if } \omega_{\text{macro}}, 22.5 < z_t \leq 100
\end{cases}
\end{equation}

where $\omega_5 = 22.4 + 35.3 \log d_{g_m} (u^{p}_{t}) + 21.3 \log f_c - 0.3 (z_t - 1.5)$, $\omega_6 = 32.4 + (43.2 - 7.6 \log z_t) \log d_{g_m} (u^{p}_{t}) + 20 \log f_c$, $\omega_7 = 13.54 + 39.08 \log d_{g_m} (u^{p}_{t}) + 20 \log f_c - 0.6 (z_t - 1.5)$, and $\omega_8 = -17.5 + (46 - 7 \log z_t) \log d_{g_m} (u^{p}_{t}) + 20 \log( \frac{40\pi f_c}{3})$.

\begin{figure}[!t]
\centering
\includegraphics[width=0.5\textwidth]{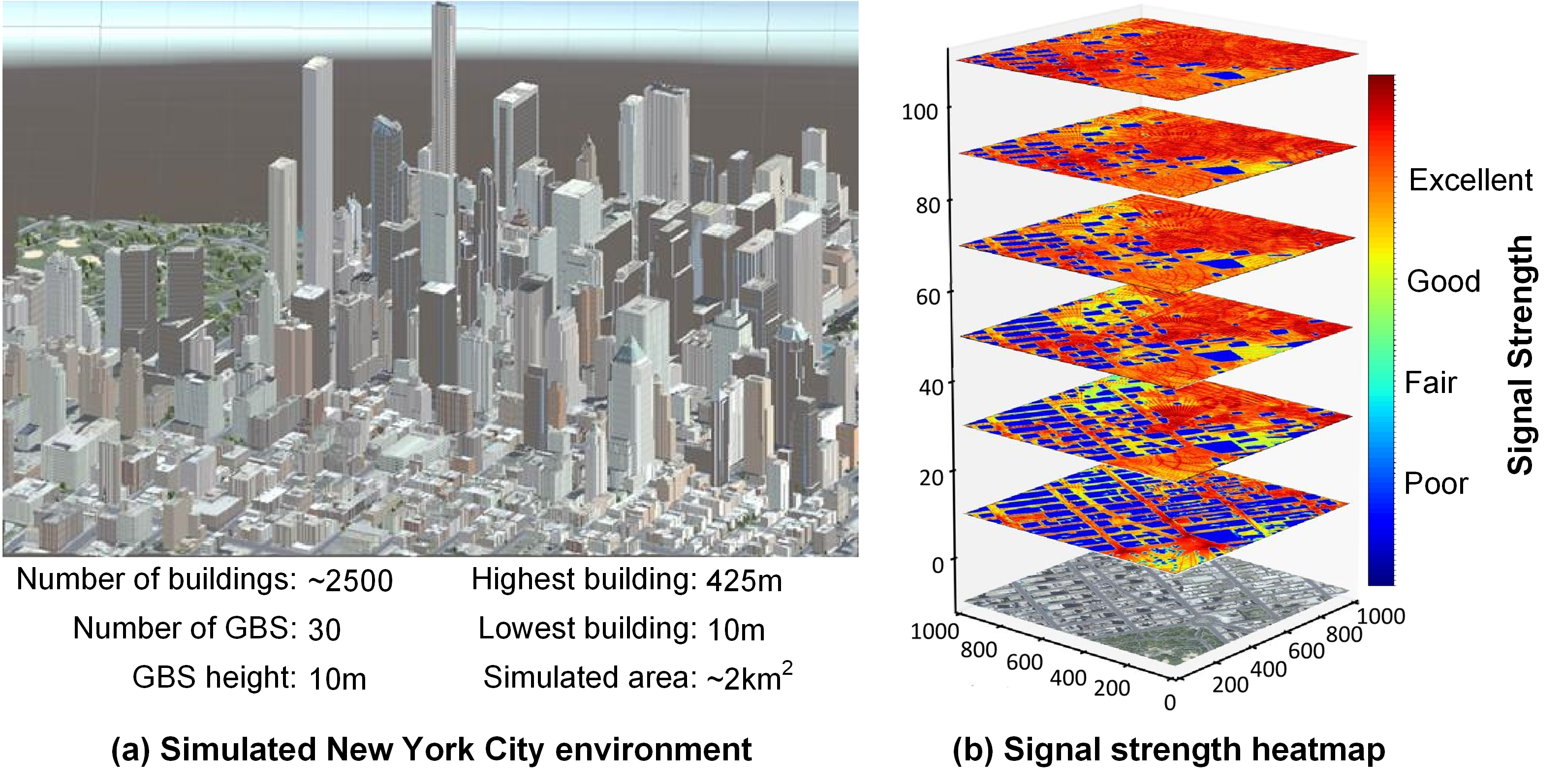}
\caption{Snapshot of the simulated 3D dense urban environment of New York City, USA, and RSRP coverage heatmap for multiple altitudes caused by 30 base Stations.}
\label{fig_6}
\end{figure}

%%%%%%%%%%%%%%%%%%%%%%%%%%%%%%%%%%%%%%%%%%%%%%%%%
                % Section III
%%%%%%%%%%%%%%%%%%%%%%%%%%%%%%%%%%%%%%%%%%%%%%%%%

\section{DRL-Driven UAV Path Planning with Assured Connectivity (DUPAC)}
The proposed DUPAC effectively designs a UAV flight path that ensures uninterrupted connectivity between the UAV and GBSs throughout its flight. It exploits DRL to design an optimal path for UAV by solving sequential decision-making problems that are formulated within the context of a Markov Decision Process (MDP). It consists of an agent that continuously interacts with its environment by collecting states, applying actions, and calculating rewards to learn an optimal decision policy. At time $t$, the agent observes environment state $s_t$, decides to perform action $a_t$, transforming the environment to next state $s_{t+1}$. The agent calculates the reward $r_{t}$ based on $s_{t+1}$ resulted from $a_t$ performed in $s_t$. The objective of the agent is to maximize the accumulative discounted reward over all time steps. 

\textbf{State}: The environment state, $s_t$, at time $t$ consists of UAV 3D position $u^{p}_{t}=(x_t,y_t,z_t)$, velocity vector $\Vec{v}_t$, the distance $c_t$ between $u^{p}_{t}$ and obstacles, the direction $\theta_t$ formed by the UAV and obstacles, and the ID of the UAV currently associated GBS $n_t$.
\begin{equation}
    s_t = (u^{p}_{t}, \Vec{v}_t, c_t, \theta_t, n_t)
\end{equation}

\textbf{Action:} The action $a_t$ is composed of 3D movement of the UAV and ID of associated GBS at $t+1$. For the 3D movement, new values for $x_{t+1}$, $y_{t+1}$, and $z_{t+1}$ are calculated by adding values between [-1,1] to $x_t$, $y_t$, and $z_t$. A GBS $g_m$ from total $M$ GBSs is selected as the associated GBS $n_{t+1}$. If $n_t$ and $n_{t+1}$ are the same then the UAV continues its association with $n_t$ otherwise it handovers to $n_{t+1}$.
\begin{equation}
    a_t = (u^{p}_{t+1}, n_{t+1})
\end{equation}

\textbf{Reward:} The reward $r_{t}$ is calculated as follows for action $a_t$ executed on the state $s_t$.
\begin{equation}
    r_t = \begin{cases}
        \mu_{1}(d_{t-1}-d_t), & \text{if } C1 \\
        \mu_{1}(d_{t-1}-d_t) + \mu_{2}w_{n_t}^{RSRP}, & \text{if } C2  ,\\
        \mu_{3}w_{n_t}^{RSRP}, & \text{otherwise}
    \end{cases}
\end{equation}
where $d_t$ is the distance from current UAV position $u^{p}_{t}$ to its destination, $w_{n_t}^{RSRP}$ is the RSRP at $u$ at $t$. $\mu_1$, $\mu_2$, and $\mu_3$ are the weights. Moreover, condition $C1$ represents excellent RSRP as (-80$\leq$RSRP), condition $C2$ represents mediocre RSRP as (-100$\leq$RSRP$<$-80), and otherwise RSRP at $u$ is poor (RSRP$<$-100). Differentiated learning of DUPAC for varying RSRP ranges allows it to distinctly control the UAV flight path for specific connectivity conditions.

This study utilizes Proximal Policy Optimization (PPO), an on-policy DRL algorithm, to learn the aforementioned DUPAC framework. Trajectory memory in PPO stores the UAV state and utilize them to train policy and value networks, as shown in Fig. 1. An advantage value $v_t$ is calculated using calculated reward $r_t$ and estimated reward from value network for the state $s_t$. Mean Squared Error (MSE) over $v_t$ is calculated to train the value network by reducing the error between estimated rewards at $s_t$ and $s_{t+1}$ with discount rate $\gamma$ \cite{R10}.

%\begin{equation}
%    D_t = \gamma V(S_{t+1})+R_t - V(S_t).
%\end{equation}
%\begin{equation}
%    L^{MSE}=\frac{1}{n} \sum_{l=0}^{n-1} D_{t+l}^{2}.
%\end{equation}

The training of policy network in PPO is conducted through a modified advantage value function $v^{GAE}$ which operates as a Generalized Advantage Estimator (GAE) to reduce the variance of estimated rewards. In addition to discount rate $\gamma$ in $v_t$, a variable $\lambda$ is included in $v^{GAE}$ for balancing a trade-off between bias and variance. In addition to $v^{GAE}$, PPO uses parameter $\epsilon$ and clip function for curtailing the variance of estimated rewards and gradually updating the policy network \cite{R10}. The clip function provides a probability ratio between $1-\epsilon$ and $1+\epsilon$, where the probability ratio is for every action by new and old stochastic policies in $s_t$. This enables the curtailment of gradient size and training of policy network with gradual learning and higher convergence rate.

\begin{figure}[!t]
\centering
\includegraphics[scale=0.4]{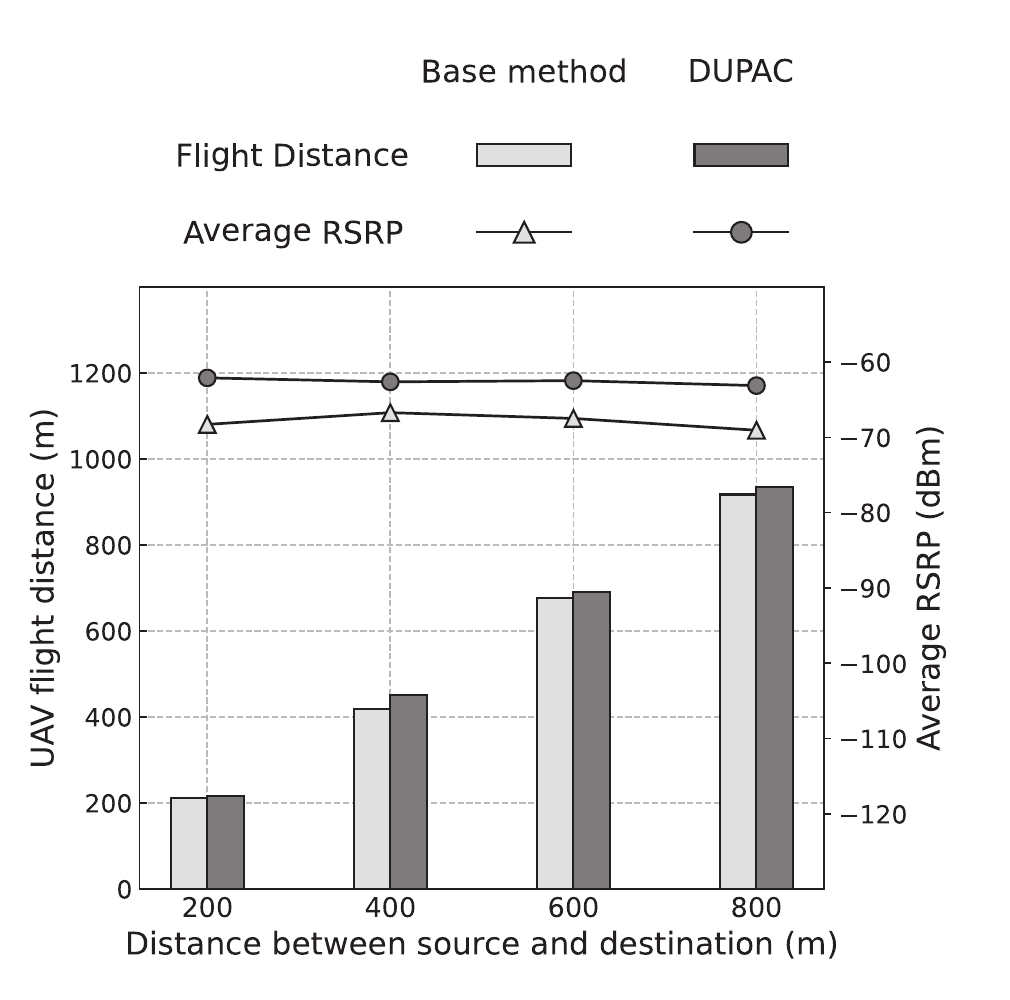}
\caption{DUPAC impact on UAV flight distance and RSRP trade-off compared to base method.}
\label{fig_3}
\end{figure}
%%%%%%%%%%%%%%%%%%%%%%%%%%%%%%%%%%%%%%%%%%%%%%%%%
                % Section IV
%%%%%%%%%%%%%%%%%%%%%%%%%%%%%%%%%%%%%%%%%%%%%%%%%
\section{Performance Evaluation}
\subsection{Experimental setup}
The evaluation of the proposed DUPAC is conducted in a dense urban environment of New York City, USA. For this purpose, a 3D simulation of $\sim$2km$^2$ area of New York City around Times Square is developed in the Unity framework, as shown in Fig. 2(a). As per real-world network deployments indicated in \cite{R11}, 30 GBS are deployed in the locations mentioned in \cite{R11}. Through the implementation of antenna gain and path loss models with 30 GBS, an RSRP coverage map is obtained for different altitudes shown in Fig. 2(b) where signal strength is poor near the ground due to congested building structures. However, the signal strength improves with the altitude as the environment is more open due to only few high rise buildings. DUPAC determines the UAV path during the flight through distance measurements to the destination and ensures the best possible connectivity via the RSRP coverage map. The evaluation of DUPAC is done for short as well as long flight distances to confirm its generalization capability for a variety of flight distances. 

The preliminary viability of DUPAC is established by comparing it with a base method. In the base method state and reward functions are the same as DUPAC but the action function only provides the next position of UAV and the selection of GBS is done via 3GPP standardized rule-based mechanism. Comparison with other state-of-the-art methods requires their implementation in our 3D simulated environment and is in the process of being included in the extended article. The values used for \( P_{g_m} \) and \( f_c \) in the path loss calculation are 15.2dBm and 2GHz, respectively. To control the UAV behaviors for different RSRP conditions, the weights $\mu_1$, $\mu_2$, and $\mu_3$ in DUPAC and base method are set as 10, 0.01, and 0.1, respectively. Moreover, the learning rate of PPO framework in DUPAC is set as 0.0003, the discount rate $\gamma$ is 0.95, and the cut-off threshold between the old and new policies is 0.2. 

\begin{figure}[!t]
\centering
\includegraphics[width=0.4\textwidth]{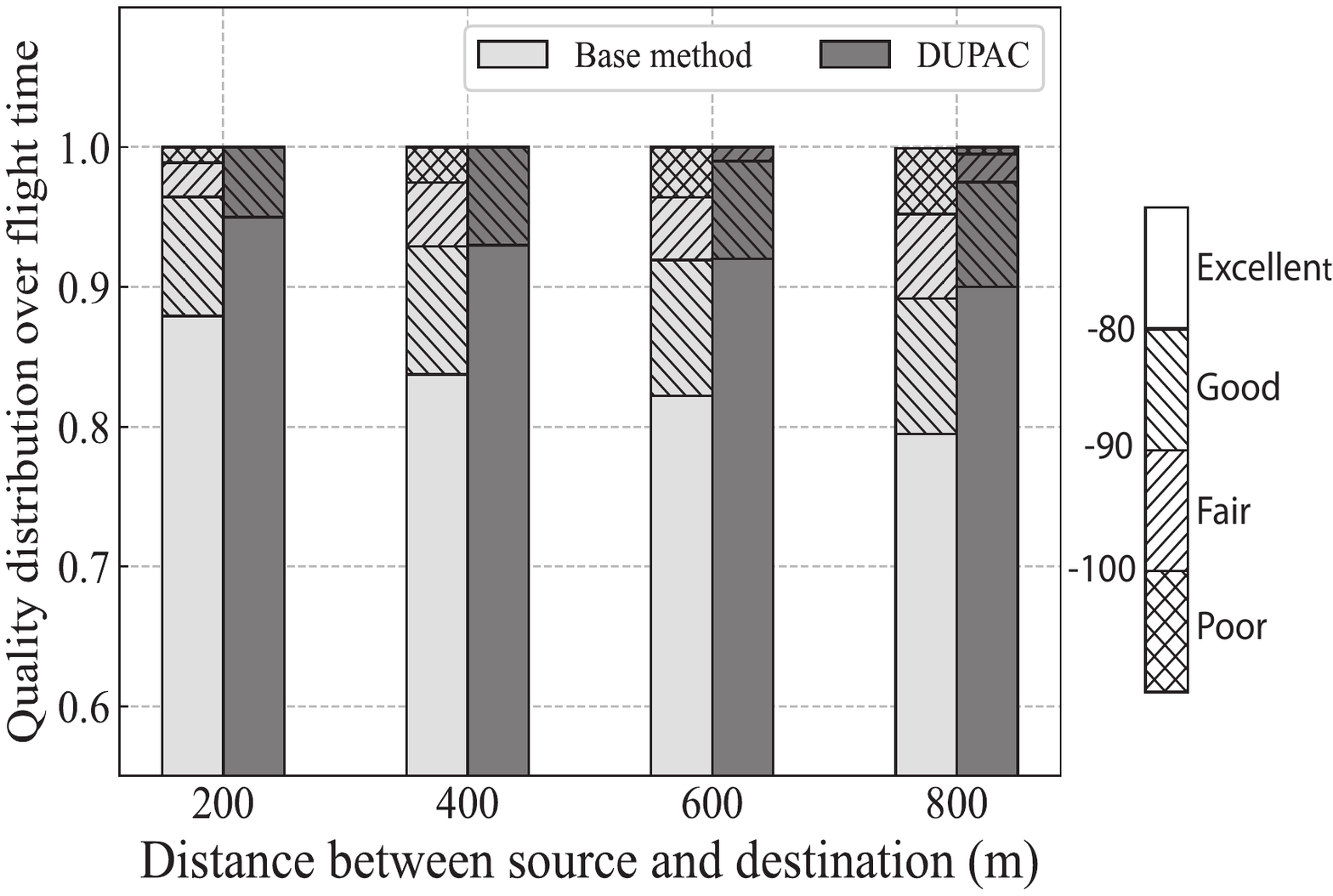}
\caption{DUPAC effect on RSRP distribution during UAV flight compared to base method.}
\label{fig_4}
\end{figure}

\subsection{Results}
The most important aspect of UAV flight planning is that UAV reaches the destination within the shortest possible distance to minimize energy consumption and elongate flight capability. To this end, UAV flight distances with DUPAC and base method are evaluated when manually calculated shortest distance between source and destination is increased from 200m to 800m. The results in Fig. \ref{fig_3} show that DUPAC covers 14\% more distance for the longest flight (800m) which is only 2\% more than the base method. On average DUPAC cover 11\% distance and base method covers 9\% more distance compared to shortest possible path to the destination. This is because DRL in the base method optimizes its policy only to minimize the distance, whereas DUPAC achieves a trade-off between the UAV flight distance and connection quality by achieving on average 7.5\% better RSRP during the UAV flight for all distances. The trade-off of distance and signal quality in DUPAC can be tweaked by adjusting the weights $\mu_1$, $\mu_2$, and $\mu_3$, and operators benefit from this flexibility by altering the primary objective of DUPAC as per service requirements.

Consistent excellent/good communication signal is critical for most of the UAV services like surveillance, videography, and rescue operations. The results in Fig. \ref{fig_4} evaluate this aspect by illustrating the distribution of RSRP at UAV throughout its flight in DUPAC and base method. It is clear that DUPAC achieves excellent RSRP for longer periods during the flight than base method. In particular, DUPAC maintains excellent RSRP on UAV for maximum 11\% and on average 9\% longer periods during the flight compared to the base method. Consequently, UAV with DUPAC has fair RSRP for a very short period during the flights for all distances, and it does not suffer with poor signal quality during flights except for 0.5\% flight duration in case of 800m distance. The enhanced signal stability from DUPAC at extended ranges improves communication reliability for UAV  and reduces communication failures. The result also reveals that as the distance between source and destination increases the flight period with excellent RSRP decreases for both DUPAC and base method, however, the rate of decrement is more for base method than DUPAC. From these results it can be concluded that the distance and signal quality trade-off policy in DUPAC successfully achieves a balance between flight distance and RSRP as it maintains excellent RSRP at UAV for longer periods at the cost of slight increase in flight distance compared to base method. 

\section{Conclusion and Future Directions}
This paper presents a DUPAC framework that uses on-policy DRL algorithm for achieving a trade-off between UAV path planning and signal quality for uninterrupted high quality communication. A generic 3D dense urban environment implemented in the Unity framework with network modeling enables evaluation of DUPAC in realistic settings and sets foundation for other UAV related studies. The comparative analysis with the base method confirms the efficacy of DUPAC in achieving the trade-off between UAV path planning and signal quality, as it achieves on average 7.5\% better signal quality with only 2\% increment in flight distance. The differentiated reward function for varying RSRP ranges enables DUPAC to keep UAV in excellent RSRP regions for on average 9\% longer period during the flights than the base method. This leads to improved data transmission quality and lower signal loss risk, thereby enhancing the reliability for mission-critical UAV operations. The in-depth evaluation of DUPAC for diverse urban environments is on-going to be included in the extended article along with the comparison with the state-of-the-art. Moreover, we are adding handover reduction functionality in DUPAC to minimize connectivity disruptions.

\bibliographystyle{IEEEtran}
\bibliography{refs}

\begin{thebibliography}{10}
\providecommand{\url}[1]{#1}
\csname url@samestyle\endcsname
\providecommand{\newblock}{\relax}
\providecommand{\bibinfo}[2]{#2}
\providecommand{\BIBentrySTDinterwordspacing}{\spaceskip=0pt\relax}
\providecommand{\BIBentryALTinterwordstretchfactor}{4}
\providecommand{\BIBentryALTinterwordspacing}{\spaceskip=\fontdimen2\font plus
\BIBentryALTinterwordstretchfactor\fontdimen3\font minus \fontdimen4\font\relax}
\providecommand{\BIBforeignlanguage}[2]{{%
\expandafter\ifx\csname l@#1\endcsname\relax
\typeout{** WARNING: IEEEtran.bst: No hyphenation pattern has been}%
\typeout{** loaded for the language `#1'. Using the pattern for}%
\typeout{** the default language instead.}%
\else
\language=\csname l@#1\endcsname
\fi
#2}}
\providecommand{\BIBdecl}{\relax}
\BIBdecl

\bibitem{R1_1}
F.~Giones and A.~Brem, ``From toys to tools: The co-evolution of technological and entrepreneurial developments in the drone industry,'' \emph{Business Horizons}, vol.~60, no.~6, pp. 875--884, 2017.

\bibitem{R1_2}
N.-N. Dao, Q.-V. Pham, N.~H. Tu, T.~T. Thanh, V.~N.~Q. Bao, D.~S. Lakew, and S.~Cho, ``Survey on aerial radio access networks: Toward a comprehensive 6g access infrastructure,'' \emph{IEEE Communications Surveys \& Tutorials}, vol.~23, no.~2, pp. 1193--1225, 2021.

\bibitem{R1_3}
G.~Geraci, A.~Garcia-Rodriguez, M.~M. Azari, A.~Lozano, M.~Mezzavilla, S.~Chatzinotas, Y.~Chen, S.~Rangan, and M.~Di~Renzo, ``What will the future of uav cellular communications be? a flight from 5g to 6g,'' \emph{IEEE communications surveys \& tutorials}, vol.~24, no.~3, pp. 1304--1335, 2022.

\bibitem{R1_4}
L.~Zhu, J.~Zhang, Z.~Xiao, X.~Cao, X.-G. Xia, and R.~Schober, ``Millimeter-wave full-duplex uav relay: Joint positioning, beamforming, and power control,'' \emph{IEEE Journal on Selected Areas in Communications}, vol.~38, no.~9, pp. 2057--2073, 2020.

\bibitem{R1_5}
Y.~Zeng, J.~Lyu, and R.~Zhang, ``Cellular-connected uav: Potential, challenges, and promising technologies,'' \emph{IEEE Wireless Communications}, vol.~26, no.~1, pp. 120--127, 2018.

\bibitem{R2}
H.~Yang, J.~Zhang, S.~Song, and K.~B. Lataief, ``Connectivity-aware uav path planning with aerial coverage maps,'' in \emph{IEEE Wireless Communications and Networking Conference (WCNC)}, 2019, pp. 1--6.

\bibitem{R4}
S.~De~Bast, E.~Vinogradov, and S.~Pollin, ``Cellular coverage-aware path planning for uavs,'' in \emph{2019 IEEE 20th International Workshop on Signal Processing Advances in Wireless Communications (SPAWC)}, 2019, pp. 1--5.

\bibitem{R5}
G.~Fontanesi, A.~Zhu, M.~Arvaneh, and H.~Ahmadi, ``A transfer learning approach for uav path design with connectivity outage constraint,'' \emph{IEEE Internet of Things Journal}, vol.~10, no.~6, pp. 4998--5012, 2022.

\bibitem{R6}
H.~Xie, D.~Yang, L.~Xiao, and J.~Lyu, ``Connectivity-aware 3d uav path design with deep reinforcement learning,'' \emph{IEEE Transactions on Vehicular Technology}, vol.~70, no.~12, pp. 13\,022--13\,034, 2021.

\bibitem{R7}
L.~F. Dorr, ``Federal aviation administration (faa) small unmanned aircraft rule part 107,'' 2016.

\bibitem{R11}
\BIBentryALTinterwordspacing
Cellmapper. (2023) Cellular network coverage maps. [Online]. Available: \url{https://www.cellmapper.net/map?MCC=311&MNC=480&type=LTE&latitude=40.76375609047827&longitude=-73.97709096997261&zoom=17.798046841204}
\BIBentrySTDinterwordspacing

\bibitem{R9}
3GPP, ``Radio access network: Study on enhanced lte support for aerial vehicles,'' \emph{Technical Specification (TS) 36.777}, 2017.

\bibitem{R10}
Y.~Jang, S.~M. Raza, M.~Kim, and H.~Choo, ``Proactive handover decision for uavs with deep reinforcement learning,'' \emph{Sensors}, vol.~22, no.~3, 2022.

\end{thebibliography}
\end{document}